\pdfoutput=1

\documentclass[11pt]{article}

\usepackage[preprint]{acl}

\usepackage{times}
\usepackage{latexsym}

\usepackage[T1]{fontenc}

\usepackage[utf8]{inputenc}

\usepackage{microtype}

\usepackage{inconsolata}

\usepackage{graphicx}
\usepackage{booktabs}
\usepackage{multicol}
\usepackage{multirow}
\usepackage{amsmath}
\usepackage{bbm}
\usepackage{cleveref}
\usepackage{wrapfig}
\usepackage{placeins}
\usepackage{subcaption}
\usepackage{caption}
\usepackage{array}
\usepackage{amssymb}
\usepackage{mathrsfs}
\usepackage[ruled]{algorithm2e}
\usepackage{xspace}

\usepackage{soul} %
\usepackage{xcolor} %
\definecolor{verylightgray}{rgb}{0.9, 0.9, 0.9}
\definecolor{verylightblue}{rgb}{0.9, 0.95, 1.0}
\definecolor{verylightpurple}{rgb}{0.95, 0.9, 1.0}
\newcommand{\hlgray}[1]{\sethlcolor{verylightgray}\hl{#1}}

\title{\logo \ \ SIFT: Grounding LLM Reasoning in Contexts via Stickers}

\newcommand{\logo}{\raisebox{-2pt}{\includegraphics[height=21pt]{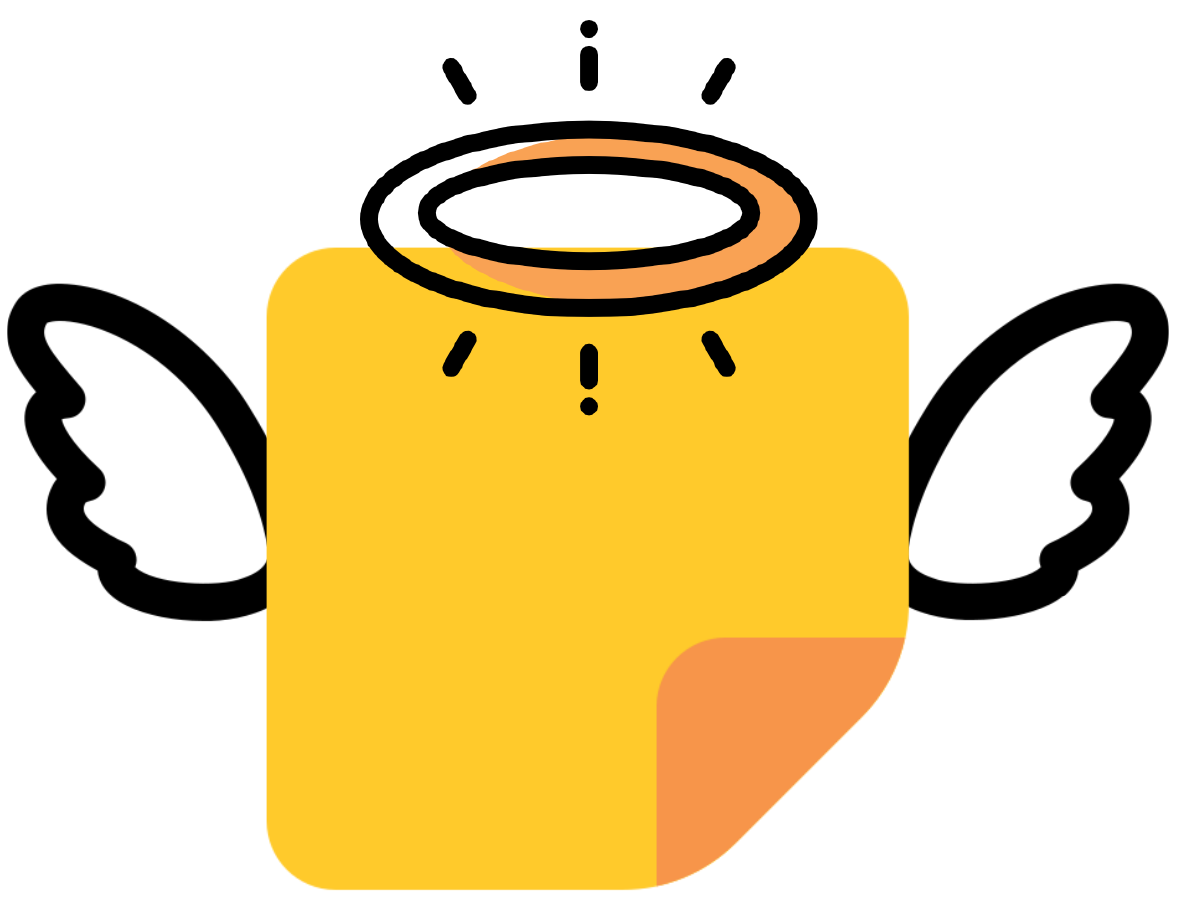}}}
\def\method{Stick to the Facts\xspace}
\def\abbr{SIFT\xspace}

\author{
    Zihao Zeng,
    Xuyao Huang\thanks{Equal contribution.},
    Boxiu Li\footnotemark[1],
    Zhijie Deng\thanks{Corresponding author.} \\
        Shanghai Jiao Tong University \\
        \{zengzihao,\;huangxuyao,\;lbxhaixing154, \;zhijied\}@sjtu.edu.cn
        }

\begin{document}
\maketitle
\begin{abstract}

This paper identifies the misinterpretation of  the context can be a significant issue during the reasoning process of large language models, spanning from smaller models like Llama3.2-3B-Instruct to cutting-edge ones like DeepSeek-R1. 
For example, in the phrase ``10 dollars per kilo,'' LLMs might not recognize that ``per'' means ``for each,'' leading to calculation errors.
We introduce a novel, post-training approach called \textbf{\underline{S}t\underline{i}ck to the \underline{F}ac\underline{t}s (\abbr)} to tackle this.
\abbr leverages increasing inference-time compute to ground LLM reasoning in contexts. 
At the core of \abbr lies the \emph{Sticker}, which is generated by the model itself to explicitly emphasize the key information within the context. 
Given the curated Sticker, \abbr generates two predictions---one from the original query and one from the query augmented with the Sticker. 
If they differ, the Sticker is sequentially refined via \emph{forward} optimization (to better align the extracted facts with the query) and \emph{inverse} generation (to conform with the model’s inherent tendencies) for more faithful reasoning outcomes. 
Studies across diverse models (from 3B to 100B+) and benchmarks (e.g., GSM8K, MATH-500) reveal consistent performance improvements.
Notably, \abbr improves the pass@1 accuracy of DeepSeek-R1 on AIME2024 from 78.33\% to \textbf{85.67}\%, establishing a new state-of-the-art in the open-source community. 
The code is available at \url{https://github.com/zhijie-group/SIFT}. 

\end{abstract}

\section{Introduction}
\label{sec:intro}

Recent advancements in large language models (LLMs)~\citep{dubey2024llama, yang2024qwen2, liu2024deepseek} have significantly advanced the field of natural language processing. 
Techniques including Chain-of-Thought (CoT) Prompting~\citep{wei2022chain, kojima2022large} and Self-Consistency~\citep{wangself}, 
as well as reasoning-enhanced models, e.g., OpenAI-o1~\citep{jaech2024openai}, DeepSeek-R1~\citep{guo2025deepseek}, and KIMI-k1.5~\citep{team2025kimi}, 
have all contributed to improvements in multi-step reasoning for solving hard problems.

Recent discussions in the community suggest that advanced reasoning capabilities in LLMs mainly stem from two factors: (i) foundational knowledge acquisition through massive pretraining on diverse data~\cite{dubey2024llama,lin2025rho1}, and (ii) strategic refinement via post-training interventions like supervised fine-tuning (SFT)~\cite{chung2022} or reinforcement learning (RL)~\cite{guo2025deepseek}, which optimize the model’s ability to select contextually relevant reasoning pathways. 
However, our studies reveal a critical lacuna in this framework: LLMs of varying sizes systematically misinterpret, overlook, or hallucinate key information in the query context---an emergent vulnerability we term \emph{factual drift}. 
\begin{figure*}[t]
    \centering
    \includegraphics[width=\linewidth]{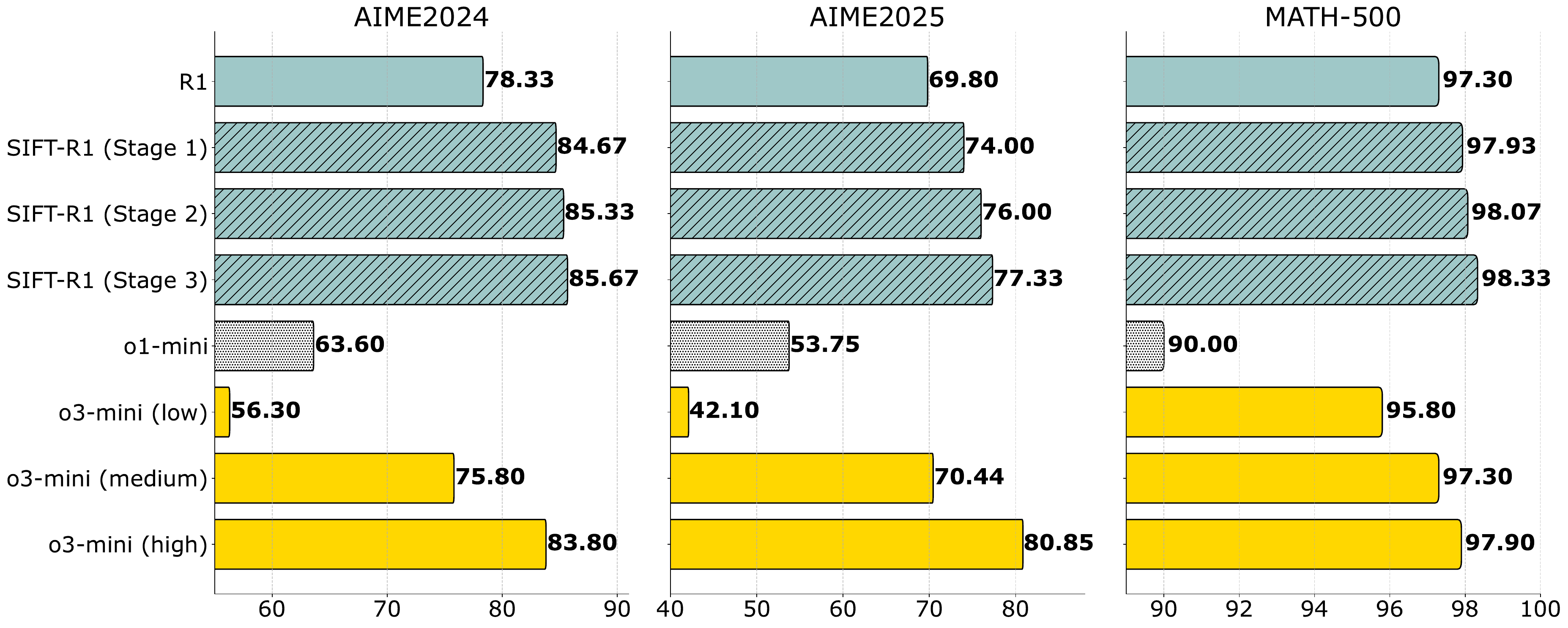}
    \caption{Applying \abbr to DeepSeek-R1 demonstrates highly competitive reasoning performance on AIME2024, AIME2025, and MATH-500 (pass@1 accuracy). The results for o1-mini and o3-mini on AIME are referenced from \citet{ye2025aimepreview}.}
    \label{fig:main1}
\end{figure*}
For example, Llama3.2-3B-Instruct~\citep{dubey2024llama} might incorrectly interpret ``per'' as ``total'' instead of ``for each'' in the phrase ``10 dollars per kilo,'' leading to reasoning errors even with the logical steps being correct. 
As a result, while current research prioritizes \emph{optimizing reasoning mechanisms} in LLMs~\cite{zelikman2022star,zelikman2024quiet,wu2024,zhang2024}, we argue equal attention should also be placed on \emph{whether LLMs are reasoning about the correct problem}. 

\begin{figure}[t]
    \centering
    \includegraphics[width=\linewidth]{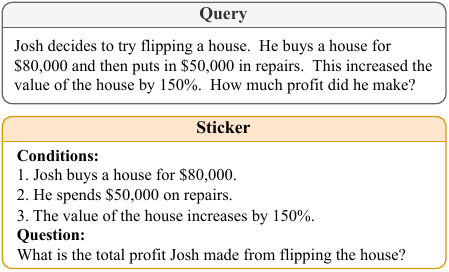}
    \caption{An example of a query and its Sticker.}
    \label{fig:Sticker_ex}
\end{figure}

We note that advanced reasoning models, such as DeepSeek-R1~\citep{guo2025deepseek}, can partially mitigate factual drift during its reasoning process via \emph{self-verification}. 
For example, the model dynamically paraphrases critical constraints (e.g., converting ``at least 3 days'' to ``minimum duration $\ge$72 hours'') to implicitly perform error-checking. 
This helps correct prior misunderstandings of the context and leads to better-aligned reasoning results.
However, such self-verification operates as a stochastic safeguard rather than a systematic protocol---it is not guaranteed to be triggered in various reasoning scenarios.
Namely, the risk of \emph{factual drift} remains and it can be significant considering the results in \Cref{fig:main1}.

Inspired by that humans usually use sticky notes to externalize critical elements when handling complex tasks, we propose the \textbf{\underline{S}t\underline{i}ck to the \underline{F}ac\underline{t}s (\abbr)} method to explicitly ground LLM reasoning in contexts using Stickers generated by the model itself. 
\abbr is a post-training approach, leveraging inference-time compute to improve generation quality yet without reliance on reward models as in Best-of-N (BoN)~\cite{brown2024large,snell2024scaling} and Monte-Carlo tree search (MCTS)~\cite{qi2024mutual,zhang2025rest}. 
Concretely, \abbr lets the target LLM summarize key facts within the input query, including essential \emph{conditions} and the core \emph{question}, into a structured \emph{Sticker} (see Figure~\ref{fig:Sticker_ex}), 
and make two predictions based on the Sticker alone and the query augmented with the Sticker, respectively. 
If they differ, the Sticker is refined through bidirectional optimization---a \emph{forward} one to better align the Sticker with the query and an \emph{inverse} one to conform to the model’s reasoning preference---for more faithful reasoning.

Experiments demonstrate that \abbr can consistently improve the reasoning performance across various LLMs and benchmarks. 
Notably, for DeepSeek-R1~\citep{guo2025deepseek}, \abbr achieves a 1.03\% accuracy improvement over the vanilla CoT (97.3\%) on MATH-500~\citep{lightman2023lets}. Additionally, on AIME2024~\citep{aime2024}, it brings a significant 7.34\% accuracy improvement  (see \Cref{fig:main1}), establishing a new state-of-the-art in the open-source community. 
We also witness a striking performance improvement for small-to-medium-sized models including  Llama3.2-3B-Instruct~\citep{dubey2024llama}, Llama3.1-8B-Instruct~\citep{dubey2024llama}, and Qwen2.5-7B-Instruct~\citep{yang2024qwen2}.

\section{Related Work}
\label{sec:related}

Reasoning has long been a significant challenge for LLMs. 
Several approaches aim to improve the reasoning capabilities of LLMs. 
These methods can be broadly categorized into techniques that align reasoning through training, enhance reasoning through search and planning, or augment reasoning during inference.

Some approaches focus on aligning the reasoning path of LLMs through Supervised Fine-Tuning (SFT) or Reinforcement Learning (RL).
STaR~\citep{zelikman2022star} enables the model to use reject sampling and learn from its mistakes by rationalizing its outputs, progressively enhancing its reasoning capabilities. 
Quiet-STaR~\citep{zelikman2024quiet} generates multiple rationales in parallel before each output token, thereby improving the model's ability to predict subsequent tokens.
V-STaR~\citep{hosseini2024v} employs a dual-system framework where the generator creates preference pairs to train the verifier, which then scores the candidate solutions.

Additionally, a significant body of work aims to enhance model reasoning abilities through search and planning. 
Q*~\citep{wang2024q} formalizes multi-step reasoning as a Markov Decision Process (MDP) and uses the A* algorithm to guide the model in selecting the optimal next step. 
rStar~\citep{qi2024mutual} employs Monte Carlo Tree Search (MCTS) to enhance the model's reasoning exploration and uses Mutual Verification to evaluate the reasoning paths. 
SR-MCTS~\citep{zhang2024llama} combines Self-Refinement and MCTS to iteratively improve and optimize newly discovered reasoning paths. 
MCTS-DPO~\citep{xie2024monte} leverages MCTS to collect step-level preference data and uses Decision-Policy Optimization (DPO) to refine the model’s policy through multiple iterations. 
ReST-MCTS*~\citep{zhang2025rest} takes a broader approach in evaluating reasoning paths, considering not only the correctness of the results but also the quality of the reasoning process, such as the shortest path and error-free intermediate steps. 
CoRe~\citep{zhu2022solving} constructs a dual-system approach with System 1 for generation and System 2 for verification, training, and reasoning simultaneously to simulate human-like reasoning processes. 
AlphaMath~\citep{chen2024alphamath} treats the output of the LLM as an action and integrates a value model and a policy model, iteratively training the model to enhance its reasoning capabilities.

\begin{figure}[t]
    \centering
    \includegraphics[width=\linewidth]{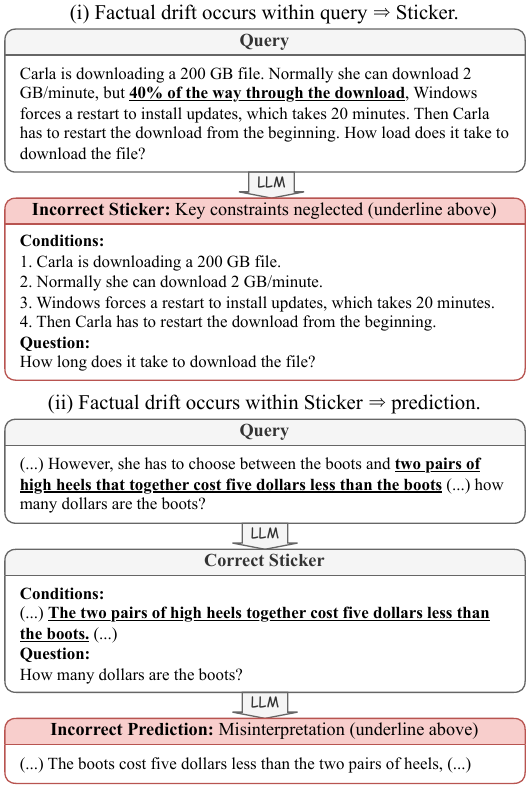}
    \caption{Factual drift occurs during (i) Sticker generation and (ii) prediction generation from Sticker.}
    \label{fig:example2}
\end{figure}

There are also methods that focus on enhancing reasoning abilities during inference. Innovations in prompt engineering have contributed to advancements in reasoning capabilities. 
Chain-of-Thought (CoT) prompting~\citep{NEURIPS2022_9d560961,kojima2022large} guides models in stepwise reasoning, such as by manually annotating natural language rationales or appending “Let’s think step by step” after questions.
Auto-CoT~\citep{zhang2022automatic} clusters questions and uses zero-shot Chain-of-Thought to generate reasoning chains, which are then used as prompts to guide the model’s answers.
ToT~\citep{yao2023tree} removes the constraints of chain structures by incorporating tree structures and search algorithms, allowing models to explore widely during reasoning. 
The seminal Self-Consistency method~\citep{wang2023selfconsistency} aggregates answers through majority voting over multiple reasoning paths, while \citet{madaan2024self} introduces iterative self-correction via feedback loops.

However, these methods primarily focus on refining \textit{how} models reason rather than ensuring that they address the \textit{correct problem}. Our approach differs by prioritizing factual comprehension before answer generation, ensuring proper problem understanding.

\section{Method}
\label{sec:method}

This section first presents the factual drift issue during LLM reasoning and then elaborates on the proposed \method (\abbr) approach. %

\begin{figure}[t]
    \centering
    \includegraphics[width=\linewidth]{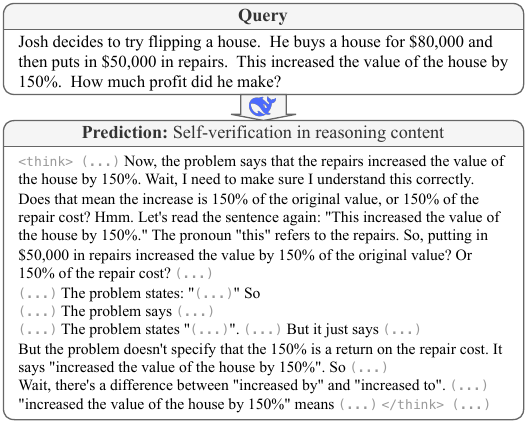}
    \caption{Self-verification occurs during DeepSeek-R1's reasoning, where the model revisiting the query, focusing on key information, and paraphrasing it.}
    \label{fig:example3}
\end{figure}

\subsection{Factual Drift in LLM Reasoning}
\label{ssec:factual}

We define \emph{factual drift} as the phenomenon where the LLM reasoning fails due to misaligned comprehension of the query context rather than flawed reasoning logic. 
This occurs when LLMs neglect key constraints, misinterpret semantic relationships, or hallucinate non-existent conditions during reasoning procedures.

We show that factual drift can be a systematic failure mode of general LLM problem-solving processes beyond reasoning. 
Taking the task of applying Stickers to Llama3.2-3B-Instruct~\citep{dubey2024llama} on GSM8K test set~\citep{cobbe2021gsm8k} as an example, we curate Stickers with the model, based on which predictions are made. 
We observe extensive factual drift errors, with typical examples displayed in \Cref{fig:example2}. 
As shown, when mapping the query to Stickers, LLMs may neglect the original constraints. %
Moreover, even when the Sticker is correct, LLMs may still misunderstand it, especially when the question is complex or uses less familiar phrasing.
The above observations also highlight that \emph{more optimization mechanisms regarding the Sticker are required to make it (i) more aligned with the query and (ii) able to be easily understood and leveraged by the target LLM}.

\noindent\textbf{Self-verification of Advanced Reasoning Models.}
We note that, for advanced models like DeepSeek-R1~\citep{guo2025deepseek}, the reasoning process sometimes involves \emph{self-verification}---revisiting the original problem, focusing on key information, and paraphrasing it. 
As illustrated in \Cref{fig:example3}, DeepSeek-R1 often states, ``Let’s read the sentence again: …'' or ``Wait, the problem states: …'' as part of its thought process, helping to deepen its understanding of the context or self-correct.

The excellent performance of such advanced reasoning models underscores the efficacy of mitigating factual drift to make the model better respect the context. 
Nevertheless, this self-verification functions more as a stochastic safeguard than a systematic protocol—it may not always be activated across different reasoning scenarios. Consequently, the risk of factual drift persists.
We consequently develop the novel SIFT framework to address this.

\begin{figure*}[t]
    \centering
    \includegraphics[width=\linewidth]{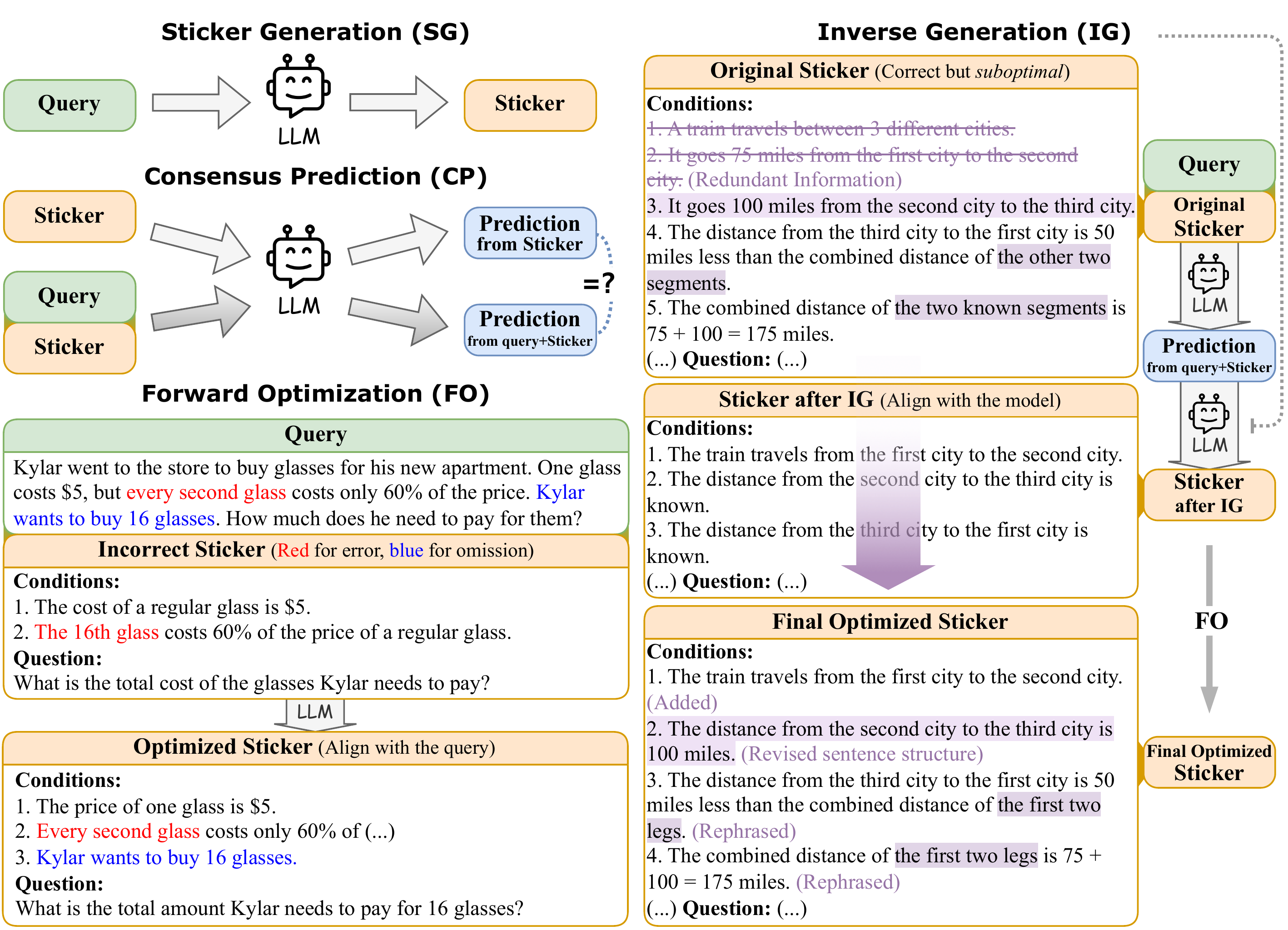}
    \caption{Four core operations in \abbr: (i) Sticker Generation (SG), (ii) Consensus Prediction (CP), (iii) Forward Optimization (FO), (iv) Inverse Generation (IG).}
    \label{fig:s2f}
\end{figure*}

\subsection{\method (\abbr)}
\label{ssec:s2f}

\abbr includes four core operations (see \Cref{fig:s2f}):  
\hlgray{(i)} Sticker Generation (SG), which extracts the Sticker from the original query;
\hlgray{(ii)} Consensus Prediction (CP), which validates the alignment between predictions from the Sticker and the query augmented with the Sticker;
\hlgray{(iii)} Forward Optimization (FO), which refines the Sticker to improve its alignment with the facts in the query;
\hlgray{(iv)} Inverse Generation (IG), which generates the Sticker based on the prediction inversely.

The full procedure of \abbr is shown in Algorithm~\ref{alg:Sticker_flow} with the details of Consensus Prediction in Algorithm~\ref{alg:cp}. All prompts used can be found in \Cref{sec:prompt}.
We explain some rationales below.

\begin{algorithm}[t]
\small 
\caption{LLM reasoning with \abbr}
\label{alg:Sticker_flow}
\SetKwInOut{Input}{Input}\SetKwInOut{Output}{Output}
\Input{Query $Q$}
\Output{Final result of $Q$}
\BlankLine
$S_1 \leftarrow \text{SG}(Q)$ \tcp*{Sticker generation}  
$P_1 \leftarrow \text{CP}(Q,S_1)$; %

\uIf{$P_1 \neq \leadsto$}{ 
    \Return $P_1$ \tcp*{Exit if consensus} 
} 
\Else{  
    \tcp{Forward} 
    $S_2 \leftarrow \text{FO}(Q,S_1)$, $P_2 \leftarrow \text{CP}(Q,S_2)$;
    
    \uIf{$P_2 \neq \leadsto$}{\Return $P_2$}  
    \Else{  
        \tcp{Inverse}  
     $S_3 \leftarrow \text{FO}(Q,\text{IG}(P_{Q, S_2}))$;
     
     $P_3 \leftarrow \text{CP}(Q,S_3)$;

            \Return $P_3$ \textbf{if} $P_3 \neq \leadsto$ \textbf{else} $\text{LLM}(Q)$
    }  
}  
\end{algorithm}

\begin{algorithm}[t]
\small 
\caption{Consensus Prediction (CP)}  
\label{alg:cp}  
\SetKwInOut{Input}{Input}\SetKwInOut{Output}{Output}  
\Input{Query $Q$, Sticker $S$}  
\Output{Prediction from $Q$ \& $S$, or $\leadsto$ (unequal)}  
\BlankLine  
$P_S \leftarrow \text{LLM}(S)$ \tcp*{Sticker-only}

$P_{Q,S} \leftarrow \text{LLM}(Q, S)$ \tcp*{Query+Sticker}  

\uIf{$\textsc{Equivalent}(P_S, P_{Q,S})$}{  
    \tcp{Consensus validation}  
    \Return $P_{Q,S}$ 
}  
\Else{  
    \Return $\leadsto$
}  
\end{algorithm}

\noindent\textbf{Consensus Prediction: Beyond Answer Aggregation.}
Traditional self-consistency methods sample diverse reasoning paths to aggregate answers~\cite {wang2023selfconsistency}, focusing on \emph{how} models reason. In contrast, our Consensus Prediction verifies \emph{whether models reason about the same problem} with dual representations:
\hlgray{(i)} the \emph{Sticker-Only} one, which forces the model to solve the problem using only the key conditions and the core question,
and \hlgray{(ii)} the \emph{Query+Sticker} one, which provides richer contexts.
This way, the model explores \emph{semantic invariance} rather than sampling diversity when reasoning about the answers. 

CP does not require sampling and operates with greedy decoding by default.
However, it remains compatible with stochastic sampling, as demonstrated in \Cref{tab:sc}. 
Besides, the CP operates only based on the current Sticker, preventing contamination from historical reasoning traces. 
As illustrated in Algorithm~\ref{alg:cp}, consensus between representations acts as a factual invariant—a necessary (though not sufficient) condition for correctness. 
This design intentionally avoids conflating factual grounding with reasoning quality assessment.

\noindent\textbf{Forward Optimization: Anchoring Stickers to Source Semantics.}
As discussed in \Cref{ssec:factual}, the SG process can also inevitably suffer from factual drift, where the original constraints are misrepresented or misunderstood. 
To address this, we combine the generated Sticker with the query to produce a refined Sticker.  
For example, it can correct misinterpretations, such as changing ``the 16th glass'' to ``every second glass'' in \Cref{fig:s2f}.

\noindent\textbf{Inverse Generation: Aligning Stickers to Model Reasoning Preference.}
It is frequently observed that for LLM reasoning, contexts with the same semantics but different presentations can yield distinct results. 
This implies that, after doing FO,
it can be beneficial to further refine the Sticker based on the LLM's reasoning process. 
Given this insight, we use the LLM to inversely infer a new Sticker given the model prediction.
We further invoke FO once again to the new Sticker to avoid factual drift. 
This step makes the Sticker respect the internal reasoning preferences of the model for representing facts, arranging conditions, or structuring questions (see \Cref{fig:s2f}).
This also helps the model recognize the difference between the Sticker from IG and the original question, to enable the model to capture overlooked information and generate a more comprehensive Sticker.

\section{Experiments}
\label{sec:exp}

\begin{figure*}[t]
    \centering
    \includegraphics[width=\linewidth]{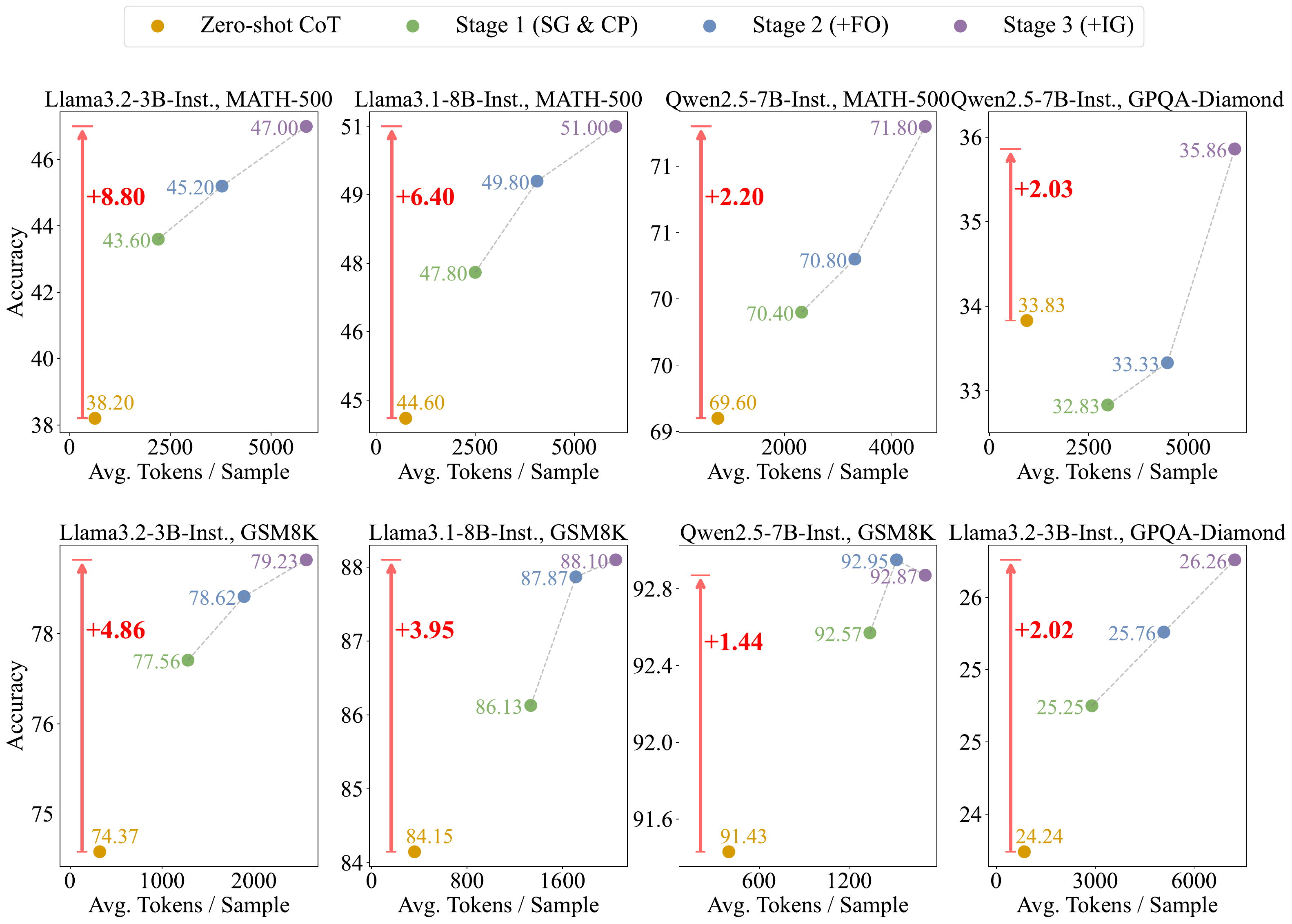}
    \caption{
    Comparison of \abbr and traditional Zero-shot CoT across multiple models and datasets.
    We divide \abbr into three stages: Stage 1 only uses SG \& CP, while Stage 2 and Stage 3 optimize the Sticker through forward (+FO) and inverse (+IG) direction, respectively. 
    The bidirectional arrows in the figure highlight the performance gap between Zero-shot CoT and the complete \abbr (i.e., Stage 3). 
    We see that in nearly all scenarios, \abbr leads to a significant performance improvement. 
    }
    \label{fig:main}
\end{figure*}

In this section, we first validate the effectiveness and generalization of \abbr (\Cref{ssec:enhance}). 
Next, we explore several variants (\Cref{ssec:iter} \& \ref{ssec:sample}). 
Finally, we include ablation studies to gain further insights into our approach (\Cref{ssec:abl}).

\subsection{Enhancing LLM Reasoning with \abbr}
\label{ssec:enhance}

\textbf{Models \& Datasets.} 
We test \abbr on a diverse set of state-of-the-art LLMs, including Llama3.2-3B-Instruct~\citep{dubey2024llama}, Llama3.1-8B-Instruct~\citep{dubey2024llama}, Qwen2.5-7B-Instruct~\citep{yang2024qwen2}, and DeepSeek-R1~\citep{guo2025deepseek}.
These models cover a range of sizes, architectures (Mixture-of-Experts (MoE) vs. dense), and reasoning capabilities.
We select well-established reasoning benchmarks, including GSM8K~\citep{cobbe2021gsm8k}, MATH-500~\citep{lightman2023lets}, GPQA-Diamond~\citep{rein2023gpqa}, and AIME2024~\citep{aime2024}.

\noindent\textbf{Test Protocol.} 
To isolate the effect of \abbr from the influence of sampling, all tests are conducted using greedy decoding, except for DeepSeek-R1. 
Because the default settings of the used Volcengine API (temperature=1.0, top-p=0.7) cannot be modified, the \abbr on DeepSeek-R1 is based on sampling. Specifically, for DeepSeek-R1 on MATH-500, we perform $3$ sampling runs and report average results. 
For AIME2024, due to its small size, we perform $10$ sampling runs and report the average. 
Additionally, we divide the entire \abbr process into three stages:
\hlgray{(i)} Stage 1: Only SG and CP are used.
\hlgray{(ii)} Stage 2: Building upon Stage 1, FO is used to optimize the Sticker.
\hlgray{(iii)} Stage 3: The complete process outlined in Algorithm~\ref{alg:Sticker_flow}.
The accuracy after each stage is measured: 
If the CP results are not aligned ($\leadsto$), the model’s direct answer to the query is used instead.
All evaluations are performed on OpenCompass~\citep{2023opencompass}.

\noindent\textbf{Main Results.}
The results are shown in \Cref{fig:main,fig:main1,fig:main2}.
As observed, \abbr consistently delivers robust and significant performance improvements compared to traditional Zero-shot CoT across all settings. 
From a methodological perspective, as the stages increase—i.e., with the forward and inverse optimization of Sticker—the average number of tokens used per sample rises, and accuracy shows an upward trend as well.
From a model standpoint, \abbr demonstrates notable effectiveness across various scales (ranging from several billion to hundreds of billions of parameters), architectures (both dense and MoE), and paradigms (traditional and reasoning models). 
Particularly noteworthy is its significant impact on DeepSeek-R1. For instance, on MATH-500, it achieves a 1.03\% absolute accuracy improvement over an already exceptionally high baseline of 97.3\%. 
On AIME2024, it also brings a substantial absolute accuracy increase of 7.34\%.
These results indicate that even for advanced reasoning models like DeepSeek-R1, sticking to the facts remains crucial for optimal performance.

\subsection{Iterative Optimization}
\label{ssec:iter}

\begin{figure}[t]
    \centering
    \includegraphics[width=.8\linewidth]{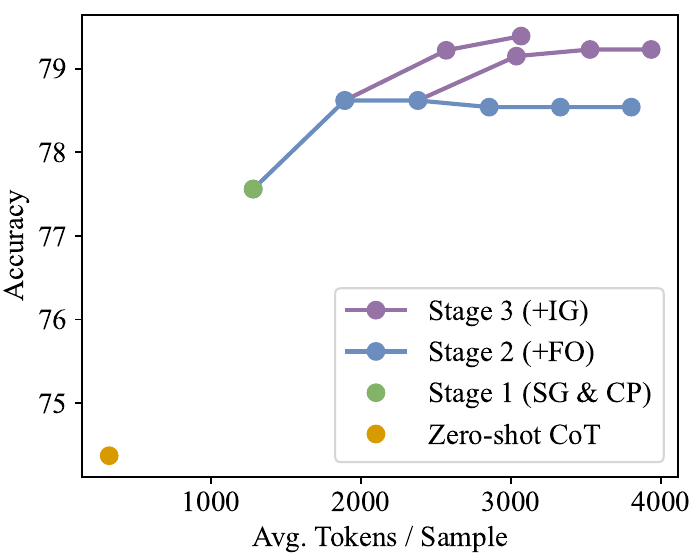}
    \caption{Iterative optimization results for \abbr. 
    The performance improves as the number of tokens per sample increases across different stages. 
    Significant gains are observed in the first repeats of Stage 2 and Stage 3.}
    \label{fig:iter}
\end{figure}

In this section, we explore whether the Sticker can be continually optimized in \abbr.

\noindent\textbf{Setup.} 
We test with Llama3.2-3B-Instruct~\citep{dubey2024llama} on the GSM8K dataset~\citep{cobbe2021gsm8k}. 
Specifically, we conduct multiple optimization repeats for Stage 2 and Stage 3. 
The other settings are the same as in \Cref{ssec:enhance}.

\noindent\textbf{Results.} 
The experimental results are shown in \Cref{fig:iter}.
We observe that \abbr shows a test-time scaling, with the performance improving as the average number of tokens per sample increases. 
For Stage 2, the saturation is rapid, but adding Stage 3 can result in an additional, noticeable performance boost. 
Nevertheless, the most significant gains are observed at the first repeat.
One possible explanation is that extracting the optimal Sticker for GSM8K is relatively easy. In more complex conditions, however, extracting a good Sticker may be harder, requiring more repeats to achieve optima.
Additionally, since we use a training-free approach for \abbr, a model trained to exclusively optimize Sticker could lead to better iterative results.

\begin{figure*}[ht]
    \centering
    \includegraphics[width=.85\linewidth]{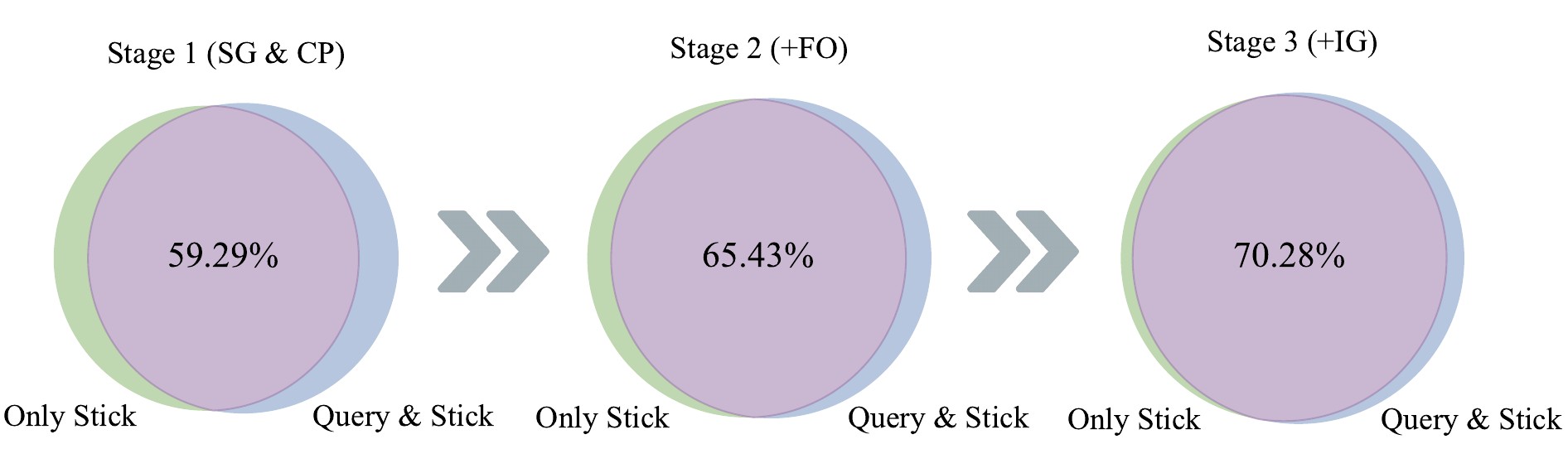}
    \caption{Venn diagrams illustrating the accuracy of predictions obtained from the ``Only Sticker'' and ``Query \& Sticker'' representations at each stage. 
    The percentages represent the accuracy where both methods correctly predict the same outcomes. 
    From Stage 1 to Stage 2, the accuracy increases by 6.14\%, and from Stage 2 to Stage 3, it increases by 4.85\%. 
    The results show the significant impact of Forward Optimization (FO) and Inverse Generation (IG) in improving prediction alignment from the two representations.}
    \label{fig:venn}
\end{figure*}

\vspace{-5pt}

\subsection{Sample Augmentation}
\label{ssec:sample}

In this section, we explore the use of Self-Consistency (SC)~\citep{wang2023selfconsistency} to enhance \abbr, demonstrating how \abbr and SC can be effectively coupled together.

Specifically, \abbr and SC can be integrated in three ways: 
\hlgray{(i) Sticker-Consistency:} Multiple Sticker samples are drawn, and consistency is applied to the predictions generated by each Sticker or by the query combined with each Sticker.
\hlgray{(ii) Prediction-Consistency:} Consistency is applied separately to predictions generated using \emph{Sticker} alone and those generated with \emph{Query + Sticker}, considering their respective samples.
\hlgray{(iii) SIFT-Consistency:} End-to-end sampling is conducted across the entire \abbr to ensure consistency.
We test Llama3.2-3B-Instruct~\citep{dubey2024llama} on GSM8K~\citep{cobbe2021gsm8k} with a temperature of 0.6, a top-p of 0.9, and 10 sampling iterations.

The results of these configurations are presented in \Cref{tab:sc}.
It is observed that our method can be combined with SC to achieve better performance. Specifically, Integrating \abbr across all dimensions results in performance improvements. Notably, \abbr-Consistency provides the most significant boost, demonstrating that the simplest sampling method---end-to-end---can lead to substantial performance gains for \abbr.

\begin{table}[t]
    \centering
    \begin{tabular}{lccc}
        \toprule
        Consistency & \multirow{2}{*}{Stage 1} & \multirow{2}{*}{Stage 2} & \multirow{2}{*}{Stage 3} \\
        Dimension & ~ & ~ & ~ \\
        \midrule
        Greedy & 77.56 & 78.62 & 79.23 \\
        (i) Sticker & 78.85 & 79.65 & 80.29 \\
        (ii) Prediction & 85.37 & 86.20 & 86.28 \\
        (iii) \abbr & --- & --- & 88.25 \\
        \bottomrule
    \end{tabular}
    \caption{Performance comparison of different consistency integration strategies for \abbr across multiple stages. The results show that integrating \abbr with Self-Consistency~\cite{wang2023selfconsistency} leads to significant performance improvements, with SIFT-Consistency achieving the highest accuracy boost.}
    \label{tab:sc}
\end{table}

\subsection{Ablation}
\label{ssec:abl}

\begin{table}[t]
    \centering
    \setlength{\tabcolsep}{3pt}
    \begin{tabular}{ccccc}
        \toprule
        \multirow{2}{*}{Model} & \multirow{2}{*}{Stage 1} & \multirow{2}{*}{Stage 2} & \multirow{2}{*}{Stage 3} & Stage 3 \\
        ~ & ~ & ~ & ~ & {\small from Stage 1} \\
        \midrule
        Llama & 77.56 & 78.62 & 79.23 & 74.07 \\
        Qwen & 92.57 & 92.95 & 92.87 & 90.90 \\
        \bottomrule
    \end{tabular}
    \caption{Performance comparison of Llama3.2-3B-Instruct and Qwen2.5-7B-Instruct on GSM8K, with and without Stage 2. 
    The results show a performance drop when skipping directly from Stage 1 to Stage 3.}
    \label{tab:jump}
\end{table}

\noindent\textbf{Evolution of Consensus Across Optimization Stages.} 
The efficacy of \abbr hinges on improving agreement between predictions derived from \emph{Sticker-only} and \emph{Query + Sticker} representations through iterative refinement. 
To quantify this alignment, We select Llama3.2-3B-Instruct~\citep{dubey2024llama} on the GSM8K dataset~\citep{cobbe2021gsm8k}.
We plot the accuracy of predictions obtained using ``Only Sticker'' and ``Query \& Sticker'' after each stage, visualized in the Venn diagram in \Cref{fig:venn}. 
As shown, both FO and IG significantly improve the alignment of the predictions from the two representations. 
Specifically, the accuracy where both methods correctly predict the same outcomes increased by 6.14\% from Stage 1 to Stage 2, and by an additional 4.85\% from Stage 2 to Stage 3.

\noindent\textbf{FO Required Before Adding IG.}
We investigate whether it is possible to skip directly from Stage 1 to Stage 3. 
We select Llama3.2-3B-Instruct and Qwen2.5-7B-Instruct on GSM8K. 
All settings remain the same as in \Cref{ssec:enhance}, except for skipping directly to Stage 3 after Stage 1. 
The results are shown in \Cref{tab:jump}.
As observed, skipping Stage 2 leads to a significant performance drop.
This indicates that during the initial optimization of Sticker, FO is essential to align Sticker with the query, followed by aligning it with model cognition. 
This is consistent with our experience, where the effectiveness of Sticker depends primarily on its correctness—ensuring no \emph{factual drift}—before considering its alignment with the model.

\begin{table}[t]
    \centering
    \begin{tabular}{lc}
        \toprule
        Strategy & Accuracy \\
        \midrule
        $P_{Q,S} \verb| if |P_{Q,S}\verb|=|P_S\verb| else | P_Q$ & 77.56 \\
        $P_S \verb| if |P_S\verb|=|P_Q\verb| else | P_{Q,S}$ & 77.02 \\
        $P_Q \verb| if |P_Q\verb|=|P_{Q,S}\verb| else | P_S$ & 76.04 \\
        \bottomrule
    \end{tabular}
    \caption{Performance comparison of various CP strategies. Here, $P_Q$, $P_S$, and $P_{Q,S}$ represent the predictions generated from query, Sticker, and query augmented with Sticker, respectively.
    The first row of the table represents the strategy used in \abbr, which is shown to be the optimal approach.}
    \label{tab:cp}
\end{table}

\begin{figure}[t]
    \centering
    \includegraphics[width=\linewidth]{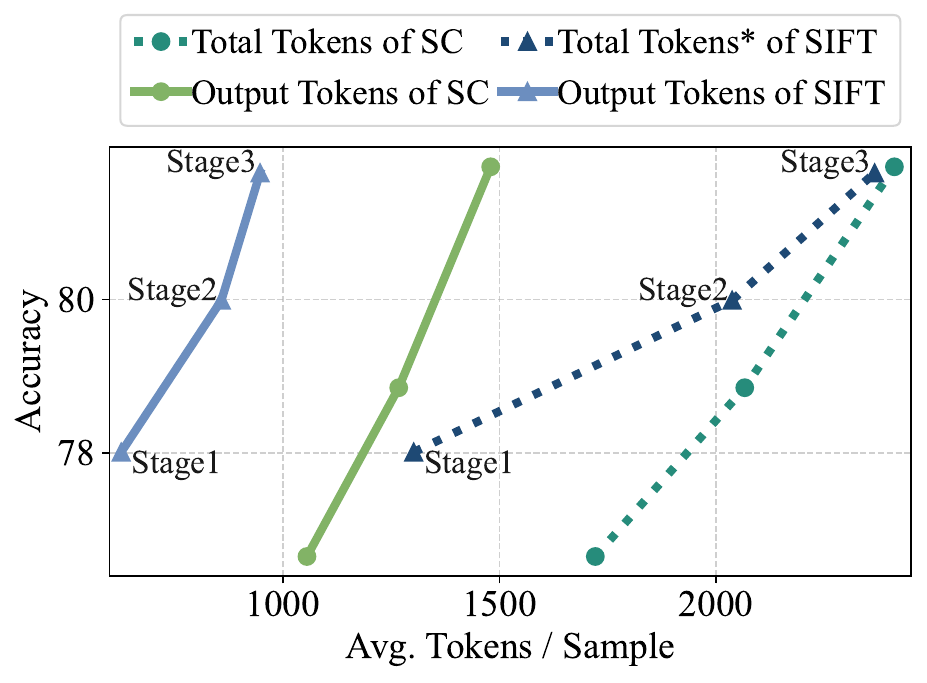}
    \caption{Comparison of \abbr and standard Self-Consistency (SC) in terms of accuracy versus average tokens per sample. The solid lines represent the output tokens used by SC (blue) and \abbr (red), while the dashed lines indicate the total tokens consumed. The ``*'' symbol in the legend denotes that the total tokens for \abbr fluctuate due to the additional formatting and example constraints used during inference. \abbr achieves comparable accuracy to SC while using significantly fewer output tokens, demonstrating its efficiency.}
    \label{fig:compare_sc2}
\end{figure}

\noindent\textbf{Comparison of \abbr and Standard Self-Consistency.}  
Under the same sampling conditions (temperature = 0.6, top-p = 0.9), we compare the performance of standard Self-Consistency (SC) with \abbr.
The evaluation is conducted using Llama3.2-3B-Instruct on GSM8K.  
For \abbr, we sample 10 times and take the average. The results are shown in \Cref{fig:compare_sc2}.  
Regarding the total tokens used by both methods, the performance curve of \abbr generally remains above that of SC. 
Notably, the ``Total Tokens*'' in the legend indicates that the total number of tokens used by \abbr varies. 
This variability arises because when generating the Sticker, we may need to provide the output format and some examples to guide its responses. 
This additional input contributes significantly to the total token count, which fluctuates considerably. We believe that further optimization can significantly reduce the total tokens required by \abbr.  
Regarding output tokens, which are more costly during inference, \abbr demonstrates a clear advantage over SC. 
Specifically, \abbr achieves a comparable performance level while using only two-thirds of the output tokens required by SC, highlighting its efficiency.

\begin{figure}[t]
    \centering
    \includegraphics[width=.9\linewidth]{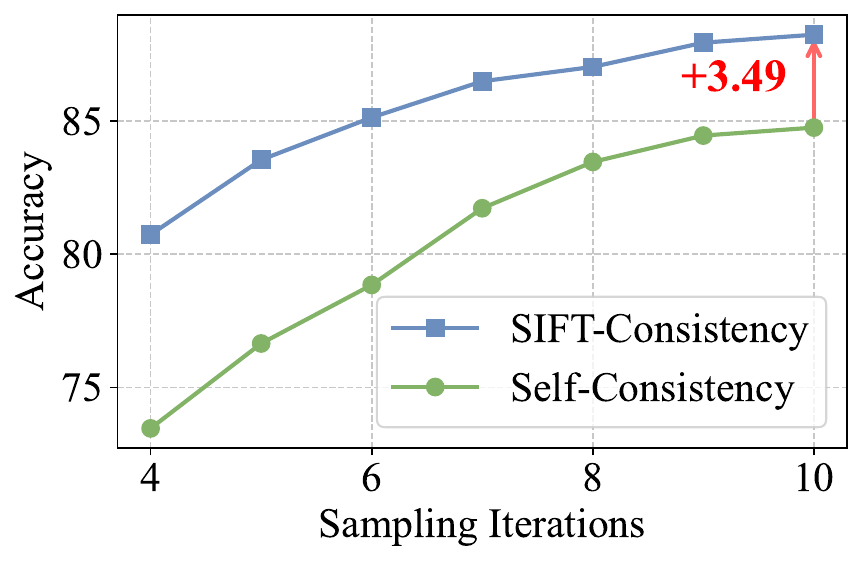}
    \caption{Comparison of SIFT-Consistency and Self-Consistency across different numbers of sampled responses per query. SIFT-Consistency consistently outperforms Self-Consistency.}
    \label{fig:compare_sc}
\end{figure}

\noindent\textbf{Comparison of SIFT-Consistency and Standard Self-Consistency.} 
In the same sampling environment (temperature = 0.6, top-p = 0.9), we compare the performance of standard Self-Consistency (SC) decoding with SIFT-Consistency, which integrates \abbr with SC. 
We conduct the evaluation using the Llama3.2-3B-Instruct model on the GSM8K dataset. 
The results are shown in \Cref{fig:compare_sc}. 
As shown in the figure, SIFT-Consistency consistently outperforms standard SC across different sampling iterations.

\noindent\textbf{Optimal Consensus Prediction Strategy.}
CP process, our strategy involves comparing predictions from \emph{Sticker} and \emph{query + Sticker}. 
If the predictions are consistent, we adopt the prediction from Query + Sticker; otherwise, we use the prediction directly from \emph{query}.
We validate this as the optimal strategy. 
Several alternative strategies were evaluated using Stage 1 results of Llama3.2-3B-Instruct on the GSM8K dataset, as shown in \Cref{tab:cp}. 
The results demonstrate that our CP strategy is effective, aligning with the prior analysis in \Cref{ssec:s2f}.

\section{Conclusion}
\label{sec:conclusion}

This study presents \method (\abbr), a training-free framework that anchors LLM reasoning to contextual facts through iterative self-refinement.
This approach enhances the reliability of LLM reasoning, providing a practical solution for factually grounded reasoning without the need for additional data or training.

\section*{Limitations}
\label{sec:lim}

This work focuses on the training-free setting. 
In the future, \abbr could be internalized into small LLMs through dedicated training, enabling more efficient on-device reasoning.
Separately, \abbr can be applied to reduce the output token length of reasoning models, improving computational efficiency without compromising accuracy. 
Additionally, Inverse Generation in \abbr offers new inspiration for data generation in inverse synthesis tasks. Further studies are needed to generalize its effectiveness across a wider range of tasks.

\bibliography{custom}

\appendix

\newpage

\section{More Results}
\label{sec:more}

\begin{figure}[ht]
\centering
\includegraphics[width=\linewidth]{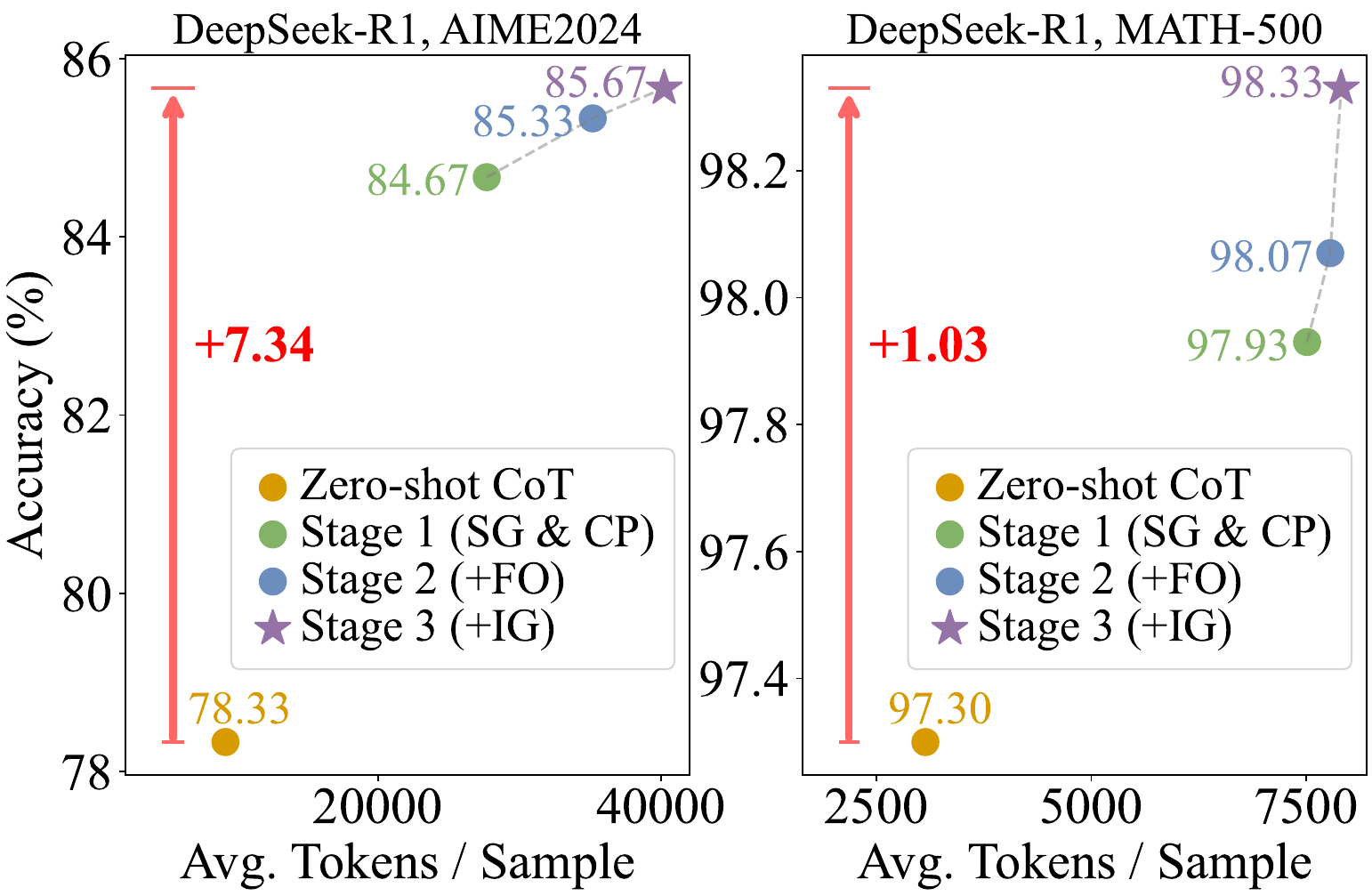}
\caption{\abbr performance on DeepSeek-R1 with increasing average token count.}
\label{fig:main2}
\end{figure}

We demonstrate how \abbr’s performance on DeepSeek-R1 evolves with an increasing average token count (see \Cref{fig:main2}).

\begin{figure}[ht]
\centering
\includegraphics[width=\linewidth]{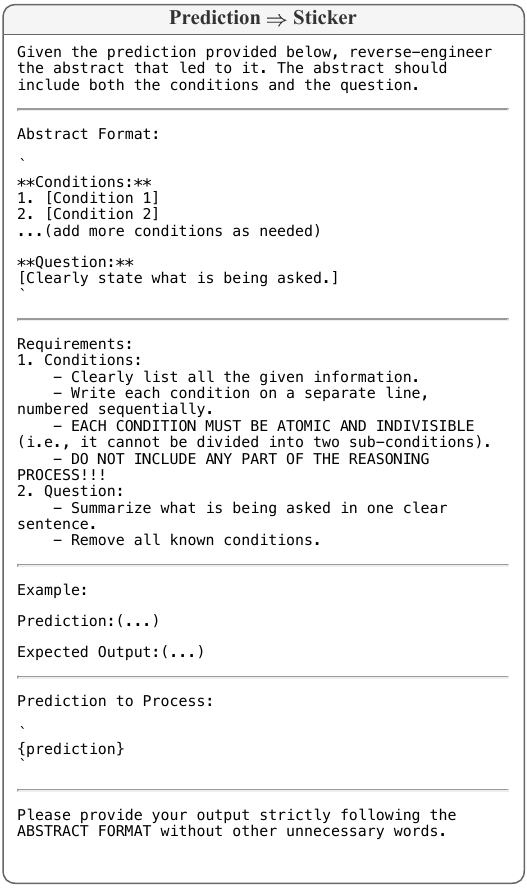}
\caption{Prompt format for generating a Sticker inversely from the prediction.}
\label{fig:p2s}
\end{figure}

\section{Prompting for \abbr}
\label{sec:prompt}

\begin{figure}[ht]
\centering
\includegraphics[width=\linewidth]{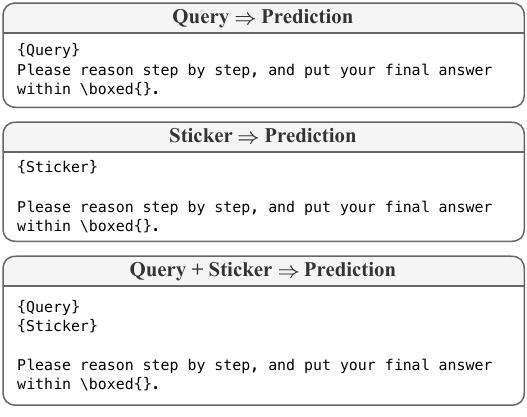}
\caption{Prompt format for generating predictions.}
\label{fig:2p}
\end{figure}

\begin{figure}[ht]
\centering
\includegraphics[width=\linewidth]{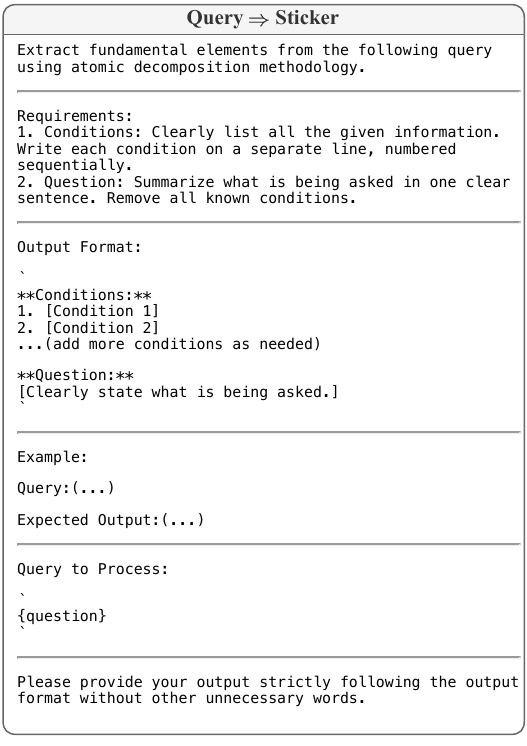}
\caption{Prompt format for generating a Sticker from the query.}
\label{fig:q2s}
\end{figure}

\begin{figure}[ht]
\centering
\includegraphics[width=\linewidth]{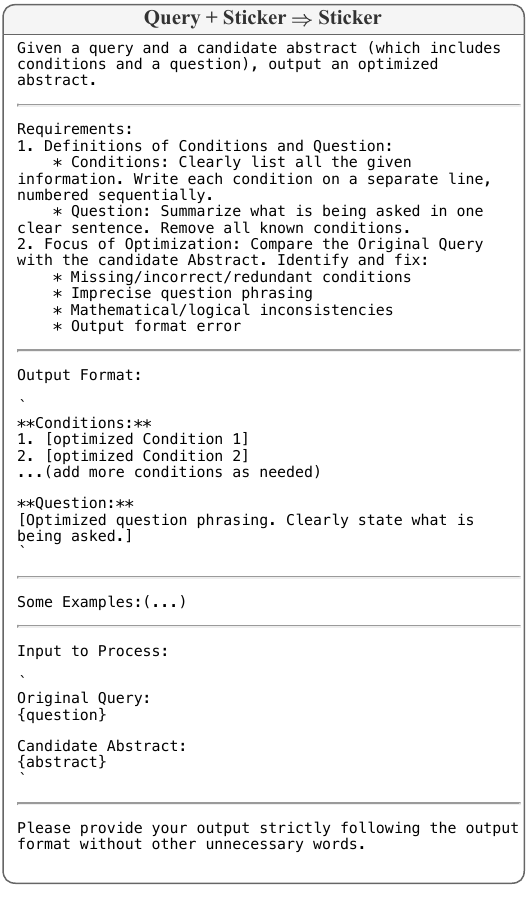}
\caption{Prompt format for forward optimization of the Sticker.}
\label{fig:q_s2s}
\end{figure}

In this section, we present the complete prompt formats used in the \abbr process (see \Cref{fig:2p,fig:q2s,fig:q_s2s,fig:p2s} for details).

\end{document}